\documentclass[manuscript, screen]{acmart}
\usepackage{booktabs} 
\usepackage{booktabs} 
\usepackage{footnote}
\usepackage{url}
\usepackage{relsize}
\usepackage{xcolor,colortbl}
\usepackage{multirow}
\usepackage{wrapfig}
\usepackage{bbm}
\usepackage[labelfont=bf]{caption}
\usepackage{tabularx}
\usepackage{amsmath,bm}
\DeclareGraphicsExtensions{.pdf,.eps,.png,.jpg,.mps}
\DeclareMathOperator\argmax{arg\,max}

\usepackage[ruled]{algorithm2e} 

\acmJournal{CIE}
\acmVolume{9}
\acmNumber{4}
\acmArticle{39}
\acmYear{2010}
\acmMonth{3}

\setcopyright{acmcopyright}

\acmDOI{0000001.0000001}

\usepackage[ruled]{algorithm2e} 

\SetAlFnt{\small}
\SetAlCapFnt{\small}
\SetAlCapNameFnt{\small}
\SetAlCapHSkip{0pt}
\IncMargin{-\parindent}

\begin{document}
\title[Two-Stream Real-time Near Accident Detection]{Intelligent Intersection: Two-Stream Convolutional Networks for Real-time Near Accident Detection in Traffic Video}

\author{Xiaohui Huang}
\orcid{1234-5678-9012-3456}
\affiliation{%
  \institution{University of Florida}
  \streetaddress{432 Newell Dr}
  \city{Gainesville}
  \state{FL}
  \postcode{32611}
  \country{USA}}
\email{xiaohuihuang@ufl.edu}

\author{Pan He}
\affiliation{%
  \institution{University of Florida}
  \streetaddress{432 Newell Dr}
  \city{Gainesville}
  \state{FL}
  \postcode{32611}
  \country{USA}}
\email{pan.he@ufl.edu}

\author{Anand Rangarajan}
\affiliation{%
  \institution{University of Florida}
  \streetaddress{432 Newell Dr}
  \city{Gainesville}
  \state{FL}
  \postcode{32611}
  \country{USA}}
\email{anand@cise.ufl.edu}

\author{Sanjay Ranka}
\affiliation{%
  \institution{University of Florida}
  \streetaddress{432 Newell Dr}
  \city{Gainesville}
  \state{FL}
  \postcode{32611}
  \country{USA}}
\email{ranka@cise.ufl.edu}

\begin{abstract}
In Intelligent Transportation System, real-time systems that monitor and analyze road users become increasingly critical as we march toward the smart city era. Vision-based frameworks for Object Detection, Multiple Object Tracking, and Traffic Near Accident Detection are important applications of Intelligent Transportation System, particularly in video surveillance and etc. Although deep neural networks have recently achieved great success in many computer vision tasks, a uniformed framework for all the three tasks is still challenging where the challenges multiply from demand for real-time performance, complex urban setting, highly dynamic traffic event, and many traffic movements. In this paper, we propose a two-stream Convolutional Network architecture that performs real-time detection, tracking, and near accident detection of road users in traffic video data. The two-stream model consists of a spatial stream network for Object Detection and a temporal stream network to leverage motion features for Multiple Object Tracking. We detect near accidents by incorporating appearance features and motion features from two-stream networks. Using aerial videos, we propose a Traffic Near Accident Dataset (TNAD) covering various types of traffic interactions that is suitable for vision-based traffic analysis tasks. Our experiments demonstrate the advantage of our framework with an overall competitive qualitative and quantitative performance at high frame rates on the TNAD dataset.
\end{abstract}

%
%

\ccsdesc[500]{Computer systems organization~Neural networks}
\ccsdesc[500]{Computer systems organization~Real-time system architecture}

%
%

\keywords{Intelligent transportation systems, near accident detection, multi-object tracking, convolutional neural networks}
\maketitle

\renewcommand{\shortauthors}{X. Huang et al.}

\section{Introduction}\label{sec:intro}
The technologies of Artificial Intelligence (AI) and Internet of Things (IoTs) are ushering in a new promising era of ''Smart Cities'', where billions of people around the world can improve the quality of their life in aspects of transportation, security, information and communications and etc. One example of the data-centric AI solutions is computer vision technologies that enables vision-based intelligence at the edge devices across multiple architectures. Sensor data from smart devices or video cameras can be analyzed immediately to provide real-time analysis for the Intelligent Transportation System (ITS). At traffic intersections, it has more volume of road users (pedestrians, vehicles), traffic movement, dynamic traffic event, near accidents and etc. It is a critically important application to enable global monitoring of traffic flow, local analysis of road users, automatic near accident detection.

As a new technology, vision-based intelligence has a wide range of applications in traffic surveillance and traffic management~\cite{coifman1998real,valera2005intelligent,buch2011review,kamijo2000traffic,veeraraghavan2003computer, he2017single}. Among them, many research works have focused on traffic data acquirement with aerial videos~\cite{angel2002methods,salvo2017traffic}, where the aerial view provides better perspectives to cover a large area and focus resources for surveillance tasks. Unmanned Aerial Vehicles (UAVs) and omnidirectional cameras can acquire useful aerial videos for traffic surveillance especially at intersections with a broader perspective of the traffic scene, with the advantage of being both mobile, and able to be present in both time and space. UAVs has been exploited in a wide range of transportation operations and planning applications including emergency vehicle guidance, track vehicle movements. A recent trend of vision-based intelligence is to apply computer vision technologies to these acquired intersection aerial videos  ~\cite{scotti2005dual,wang2006intelligent} and process them at the edge across multiple ITS architecture.  

From global monitoring of traffic flow for solving traffic congestion to quest for better traffic information, an increasing reliance of ITS has resulted in a need for better object detection (such as wide-area detectors for pedestrian, vehicles), and  multiple vehicle tracking that yields  traffic parameters such as flow, velocity and vehicle trajectories. Tracks and trajectories are measures over a length of path rather than at a single point. It is possible to tackle related surveillance tasks including  traffic movement measurements (e.g. turn movement counting) and routing information. The additional information from vehicle trajectories could be utilized to improve near accident detection, by either detecting stopped vehicles with their collision status or identifying acceleration / deceleration patterns or conflicting trajectories that are indicative of near accidents. Based on the trajectories, it is also possible to learn and forecast vehicle trajectory to enable near accident anticipation. 

Generally, a vision-based surveillance tool for intelligent transportation system should meet several requirements:
\begin{enumerate}
    \item Segment vehicles from the background and from other vehicles so that all vehicles (stopped or moving) are detected;
    \item Classify detected vehicles into categories: cars, buses, trucks, motorcycles and etc;
    \item Extract spatial and temporal features (motion, velocity, trajectory) to enable more specific tasks including vehicle tracking, trajectory analysis, near accident detection, anomaly detection and etc;
    \item Function under a wide range of traffic conditions (light traffic, congestion, varying speeds in different lanes) and a wide variety of lighting conditions (sunny, overcast, twilight, night, rainy, etc.);
    \item Operate in real-time.
\end{enumerate} Over the decades, although an increasing number of research on vision-based system for traffic surveillance have been proposed, many of these criteria still cannot be met. Early solutions~\cite{hoose1992impacts} do not identify individual vehicles as unique targets and progressively track their movements. Methods have been proposed to address individual vehicle detection and vehicles tracking problems~\cite{koller1993model,mclauchlan1997real,coifman1998real} with tracking strategies including model based tracking, region based tracking, active contour based tracking, feature based tracking and optical flow employment. Compared to traditional hand-crafted features,  deep learning methods~\cite{ren2015faster,girshick2016region,redmon2016you, tian2016detecting} in object detection have illustrated the robustness with specialization of the generic detector to a specific scene. Leuck~\cite{leuck1999automatic} and Gardner~\cite{gardner1996interactive} use three-dimensional (3-D) models of vehicle shapes to estimate vehicle images projected onto a two-dimensional (2-D) image plane. Recently, automatic traffic accident detection has become an important topic. One typical approach uses object detection or tracking before detecting accident events~\cite{sadeky2010real,kamijo2000traffic,jiansheng2014vision, jiang2007abnormal,hommes2011detection}, with  Histogram of Flow Gradient
(HFG), Hidden Markov Model (HMM) or, Gaussian Mixture Model (GMM).  Other approaches~\cite{liu2010anomaly,ihaddadene2008real,wang2010anomaly,wang2012real,tang2005traffic,karim2002incident,xia2015vision,chen2010automatic,chen2016vision} use low-level features (e.g.\ motion features) to demonstrate better robustness. Neural networks have also been employed to automatic accident detection~\cite{ohe1995method,yu2008back,srinivasan2004evaluation,ghosh2003wavelet}.

In this paper, we first propose a Traffic Near Accident Dataset (TNAD). Intersections tend to experience more and severe near accident, due to factors such as angles and turning collisions. Observing this, the TNAD dataset is collected to contain three types of video data of traffic intersections that could be utilized for not only near accident detection but also other traffic surveillance tasks including turn movement counting. The first type is drone video that monitoring an intersections with top-down view. The second type of intersection videos is real traffic videos acquired by omnidirectional fisheye cameras that monitoring small or large intersections. It is widely used in transportation surveillance. These video data can be directly used as inputs for any vision-intelligent framework. The pre-processing of fisheye correction can be applied to them for better surveillance performance. 
As there exist only a few samples of near accident in the reality per hour. The third type of video is proposed by simulating with game engine for the purpose to train and test with more near accident samples. 

We propose a uniformed vision-based framework with the two-stream Convolutional Network architecture that performs real-time detection, tracking, and near accident detection of traffic road users. The two-stream Convolutional Networks consist of a spatial stream network to detect individual vehicles and likely near accident regions at the single frame level, by capturing appearance features with a state-of-the-art object detection method~\cite{redmon2016you}. The temporal stream network leverages motion features extracted from detected candidates to perform multiple object Tracking and generate corresponding trajectories of each tracking target. We detect near accident by incorporating appearance features and motion features to compute probabilities of near accident candidate regions. Experiments demonstrate the advantage of our framework with an overall competitive performance at high frame rates. The contributions of this work can be summarized as: 
\begin{itemize}
    \item A uniformed framework that performs real-time object detection, tracking and near accident detection.
    \item The first work of an end-to-end trainable two-stream deep models to detect near accident with good accuracy.
    \item A Traffic and Near Accident Detection Dataset (TNAD) containing different types of intersection videos that would be used for several vision-based traffic analysis tasks.
\end{itemize}

The organization of the paper is as follows. Section~\ref{sec:background} describes background on Object Detection, Multiple Object Tracking and Near Accident Detection. Section~\ref{sec:method} describes the overall architecture, methodologies, and implementation of our vision-based intelligent framework. This is followed in Section~\ref{sec:experiments} by an introduction of our Traffic Near Accident Detection Dataset (TNAD) and video preprocessing techniques. Section~\ref{sec:experiments} presents a comprehensive evaluation of our approach and other state-of-the-art near accident detection methods both qualitatively and quantitatively. Section~\ref{sec:conclusion} concludes by summarizing our contributions and also discusses the scope for future work.

\section{Background}\label{sec:background}
\subsection{Object Detection}
Object detection has received significant attention and achieved striking improvements in recent years, as demonstrated in popular object detection competitions such as PASCAL
VOC detection challenge~\cite{everingham2010pascal, everingham2015pascal}, ILSVRC large scale detection challenge~\cite{russakovsky2015imagenet} and MS COCO large scale detection challenge~\cite{lin2014microsoft}. Object detection aims at outputting instances of semantic objects with a certain class label such as humans, cars. It has wide applications in many computer vision tasks including face detection, face recognition, pedestrian detection, video object co-segmentation, image retrieval, object tracking and video surveillance. Different from image classification, object detection is not to classify the whole image. 
Position and category information of the objects are both needed which means we have to segment instances of objects from backgrounds and label them with position and class. The inputs are images or video frames while the outputs are lists where each item represents position and category information of candidate objects. In general, object detection seeks to extract discriminative  features to help in distinguishing the classes.

Methods for object detection generally fall into 3 categories: 1) traditional machine learning based approaches; 2) region proposal based deep learning approaches; 3) end-to-end deep learning approaches. For traditional machine learning based approaches, one of the important steps is to design features. Many methods have been proposed to first design features~\cite{viola2001rapid,viola2004robust,lowe1999object,dalal2005histograms} and apply techniques such as support vector machine (SVM)~\cite{hearst1998support} to do the classification. The main steps of traditional machine learning based approaches are:
\begin{itemize}
    \item Region Selection: using sliding windows at different sizes to select candidate regions from whole images or video frames;
    \item Feature Extraction: extract visual features from candidate regions using techniques such as Harr feature for face detection, HOG feature for pedestrian detection or general object detection;
    \item Classifier: train and test classifier using techniques such as SVM.
\end{itemize} The tradition machine learning based approaches have their limitations. The scheme using sliding windows to select RoIs (Regions of Interests) increases computation time with a lot of window redundancies. On the other hand, these hand-crafted features are not robust due to the diversity of objects, deformation, lighting condition, background and etc., while the feature selection has a huge effect on classification performance of candidate regions. 

Recent advances in deep learning, especially in computer vision have shown that Convolutional Neural Networks (CNNs) have a strong capability of representing objects and help to boost the performance of numerous vision tasks, comparing to traditional heuristic features \cite{dalal2005histograms}. For deep learning based approaches, there are convolutional neural networks (CNN) to extract features of region proposals or end-to-end object detection without specifically defining features of a certain class. The well-performed deep learning based approaches of object detection includes Region Proposals (R-CNN)~\cite{girshick2014rich}, Fast R-CNN~\cite{girshick2015fast}, Faster R-CNN~\cite{ren2015faster}), Single Shot MultiBox Detector (SSD)~\cite{liu2016ssd}, and You Only Look Once (YOLO)~\cite{redmon2016you}.

Usually, we adopt region proposal methods (Category 2) for producing multiple object proposals, and then apply a robust classifier to further refine the generated proposals, which are also referred as two-stage method. The first work of the region proposal based deep learning approaches is R-CNN~\cite{girshick2014rich} proposed to solve the problem of selecting a huge number of regions. The main pipeline of R-CNN~\cite{girshick2014rich} is: 1) gathering input images; 2) generating a number of region proposals (e.g. 2000); 3) extracting CNN features; 4) classifying regions using SVM. It usually adopts Selective Search (SS), one of the state-of-art object proposals method \cite{Uijlings13} applied in numerous detection task on several fascinating systems\cite{girshick2014rich,girshick2015fast,ren2015faster},  to extract these regions from the image and names them region proposals. Instead of trying to classify all the possible proposals, R-CNN select a fixed set of proposals (e.g. 2000) to work with. The selective search algorithm used to generate these region proposals includes: (1) Generate initial sub-segmentation, generate many candidate regions; (2) Use greedy algorithm to recursively combine similar regions into larger ones; (3) Use the generated regions to produce the final candidate region proposals.

These candidate region proposals are warped into a square and fed into a convolutional neural network (CNN) which acts as the feature extractor. The output dense layer consists of the extracted features to be fed into an SVM~\cite{hearst1998support} to classify the presence of the object within that candidate region proposal. The main problem of R-CNN~\cite{girshick2014rich} is that it is limited by the inference speed, due to a huge amount of time spent on extracting features of each individual region proposal. And it cannot be applied in applications requiring a real-time performance (such as online video analysis). Later, Fast R-CNN~\cite{girshick2015fast} is proposed to improve the speed by avoiding feeding raw region proposals every time. Instead, the convolution operation is done only once per image and RoIs over the feature map are generated. Faster R-CNN \cite{ren2015faster} further exploits the shared convolutional features to extract region proposals used by the detector. Sharing convolutional features leads to substantially faster speed for object detection system.

The third type is end-to-end deep learning approaches which do not need region proposals (also referred as one-stage method). The pioneer works are SSD~\cite{liu2016ssd} and YOLO~\cite{redmon2016you} . An SSD detector \cite{liu2016ssd} works by adding a sequence of feature maps of progressively decreasing the spatial resolution
to replace the two stage's second classification stage, allowing a fast computation and multi-scale detection on one single input. YOLO detecor is an object detection algorithm much different from the region based algorithms. In YOLO~\cite{redmon2016you}, it regards object detection as an end-to-end regression problem and uses a single convolutional network to predict the bounding boxes and the corresponding class probabilities. It first takes the image and splits it into an $S \times S$ grid, within each of the grid we take $m$ bounding boxes. For each of the bounding box with multi scales, the convolutional neural network outputs a class probability and offset values for the bounding box. Then it selects bounding boxes which have the class probability above a threshold value and uses them to locate the object within the image. YOLO~\cite{redmon2016you} is orders of magnitude faster (45 frames per second) than other object detection approaches but the limitation is that it struggles with small objects within the image.

\subsection{Multiple Object Tracking}
Video object tracking is to locate objects over video frames and it has various important applications in robotics, video surveillance and video scene understanding. Based on the number of moving objects that we wish to track, there are Single Object Tracking (SOT) problem and Multiple Object Tracking (MOT) problem. In addition to detecting objects in video frame, the MOT solution requires to robustly associate multiple detected objects between frames to get a consistent tracking and this data association part remains very challenging. In MOT tasks, for each frame in a video, we aim at localizing and identifying all objects of interests, so that the identities are consistent throughout the video. Typically, the main challenge lies on speed, data association, appearance change, occlusions, disappear / re-enter objects and etc. In practice, it is desired that the tracking could be performed in real-time so as to run as fast as the frame-rate of the video. Also, it is challenging to provide a consistent labeling of the detected objects in complex scenarios such as objects change appearance, disappear, or involve severe occlusions.

In general, Multiple Object Tracking (MOT) can be regarded as a multi-variable estimation problem~\cite{luo2014multiple}. The objective of multiple object tracking can be modeled by performing MAP (maximal a posteriori) estimation in order to find the \textit{optimal} sequential states of all the objects, from the conditional distribution of the sequential states given all the observations:
\begin{equation}
\label{eq:map}
\widehat{\mathbf{S}}_{1:t} = \underset{\mathbf{S}_{1:t}}\argmax \ P\left(\mathbf{S}_{1:t}|\mathbf{O}_{1:t}\right).
\end{equation} where $\mathbf{s}_t^i$ denotes the state of the $i$-th object in the $t$-th frame. $\mathbf{S}_t = (\mathbf{s}_t^1, \mathbf{s}_t^2, ..., \mathbf{s}_t^{M_t})$ denotes states of all the $M_t$ objects in the $t$-th frame. $\mathbf{S}_{1:t} = \{\mathbf{S}_1, \mathbf{S}_2, ..., \mathbf{S}_t\}$ denotes all the sequential states of all the objects from the first frame to the $t$-th frame. In tracking-by-detection, $\mathbf{o}_t^i$ denotes the collected observations for the $i$-th object in the $t$-th frame. $\mathbf{O}_t = (\mathbf{o}_t^1, \mathbf{o}_t^2, ..., \mathbf{o}_t^{M_t})$ denotes the collected observations for all the $M_t$ objects in the $t$-th frame. $\mathbf{O}_{1:t} = \{\mathbf{O}_1, \mathbf{O}_2, ..., \mathbf{O}_t\}$ denotes all the collected sequential observations of all the objects from the first frame to the $t$-th frame. Different Multiple Object Tracking (MOT) algorithms can be thought as designing different approaches to solving the above MAP problem, either from a \emph{probabilistic inference} perspective, e.g. Kalman filter or a \emph{deterministic optimization} perspective, e.g. Bipartite graph matching, and machine learning approaches.

Multiple Object Tracking (MOT) approaches can be categorized by different types of models. A distinction based on \textit{Initialization Method} is that of Detection Based Tracking (DBT) versus Detection Free Tracking (DFT). DBT refers that before tracking, object detection is performed on video frames. DBT methods involve two distinct jobs between the detection and tracking of objects. In this paper, we focus on DBT, also refers as tracking-by-detection for MOT. The reason is that DBT methods are widely used due to excellent performance with deep learning based object detectors, while DFT methods require manually annotations of the targets and bad results could arise when a new unseen object appears. Another important distinction based on \textit{Processing Mode} is that of Online versus Offline models. An Online model receives video input on a frame-by-frame basis, and gives output per frame. This means only information from past frames and the current frame can be used. Offline models have access to the entire video, which means that information from both past and future frames can be used.

Tracking-by-detection methods are usually utilized in online tracking models. A simple and classic pipeline is as (1) Detect objects of interest; (2) Predict new locations of objects from previous frames; (3) Associate objects between frames by similarity of detected and predicted locations. Well-performed CNN architectures can be used for object detection such as Faster R-CNN~\cite{ren2015faster}, YOLO~\cite{redmon2016you} and SSD~\cite{liu2016ssd}. For prediction of new locations of tracked objects, approaches model the velocity of objects, and predict the position in future frames using optical flow, or recurrent neural networks, or Kalman filters. The association task is to determine which detection corresponds to which object, or a detection represents a new object. 

One popular dataset for Multiple Object Tracking (MOT) is MOTChallenge~\cite{leal2015motchallenge}. In MOTChallenge~\cite{leal2015motchallenge}, detections for each frame are provided in the dataset, and the tracking capability is measured as opposed to the detection quality. Video sequences are labeled with bounding boxes for each pedestrian collected from multiple sources. This motivates the use of tracking-by-detection paradigm. MDPs~\cite{xiang2015learning} is a tracking-by-detection method and achieved the state-of-the-art performance on MOTChallenge~\cite{leal2015motchallenge} Benchmark when it was proposed. Major contributions can be solving MOT by learning a MDP policy in a reinforcement learning fashion which benefits from both advantages of offline-learning and online-learning for data association. It also can handle the birth / death and appearance / disappearance of targets by simply treating them as state transitions in the MDP while leveraging existing online single object tracking methods. SORT~\cite{bewley2016simple} is a simple and real-time Multiple Object Tracking (MOT) method where state-of-the-art tracking quality can be achieved with only classical tracking methods. It is the most widely used real-time online Multiple Object Tracking (MOT) method and is very efficient for real-time applications in practice. Due to the simplicity of SORT~\cite{bewley2016simple}, the tracker updates at a rate of 260 Hz which is over 20x faster than other state-of-the-art trackers. On the MOTChallenge~\cite{leal2015motchallenge}, SORT~\cite{bewley2016simple} with a state-of-the-art people detector ranks on average higher than MHT~\cite{kim2015multiple} on standard detections. DeepSort~\cite{wojke2017simple} is an extension of SORT~\cite{bewley2016simple} which integrates appearance information to improve the performance of SORT~\cite{bewley2016simple} which can track through longer periods of occlusion, making SORT~\cite{bewley2016simple} a strong competitor to state-of-the-art online tracking algorithms.

\subsection{Near Accident Detection}
In addition to vehicle detection and vehicle tracking, analysis of the interactions or behavior of tracked vehicles has emerged as an active and challenging research area in recent years~\cite{sivaraman2011learning,hermes2009long,wiest2012probabilistic}. Near Accident Detection is one of the highest levels of semantic interpretation in characterizing the interactions of vehicles on the road. The basic task of near accident detection is to locate near accident regions and report them over video frames. In order to detect near accident on traffic scenes, robust vehicle detection and vehicle tracking are the prerequisite tasks.

Most of the near accident detection approaches are based on motion cues and trajectories. The most typical motion cues are optical flow and trajectory. Optical flow is widely utilized in video processing tasks such as video segmentation~\cite{huang2018supervoxel}. A trajectory is defined as a data sequence containing several concatenated state vectors from tracking, an indexed sequence of positions and velocities over a given time window. In recent years, researches have tried to make long-term classification and prediction of vehicle motion. Based on vehicle tracking algorithms such as Kalman filtering, optimal estimation of the vehicle state can be computed one frame ahead of time. Trajectory modeling approaches try to predict vehicle motion more frames ahead of time, based on models of typical vehicle trajectories~\cite{sivaraman2011learning,hermes2009long,wiest2012probabilistic}. In~\cite{sivaraman2011learning}, it used clustering to model the typical trajectories in highway driving and  hidden Markov modeling for classification. In~\cite{hermes2009long}, trajectories are classified using a rotation-invariant version of the longest common subsequence as the similarity metric between trajectories. In~\cite{wiest2012probabilistic}, it uses variational Gaussian mixture modeling to classify and predict the long-term trajectories of vehicles.

Over the past two decades, for automatic traffic accident detection, a great deal of literature emerged in various ways. Several approaches have been developed based on decision trees, Kalman filters, or time series analysis, with varying degrees of success in their performance~\cite{srinivasan2003traffic,srinivasan2001hybrid,xu1998real,shuming2002traffic,jiansheng2014vision,bhonsle2000database}. Ohe et al.~\cite{ohe1995method} use neural networks to detect traffic incidents immediately by utilizing one minute average traffic data as input, and determine whether an incident has occurred or not. In~\cite{ikeda1999abnormal}, the authors propose a system to distinguish between different types of incidents for automatic incident detection. In~\cite{kimachi1994incident}, it investigates the abnormal behavior of vehicle related to accident based on the concepts of fuzzy theory where accident occurrence relies on the behavioral abnormality of multiple continual images. Zeng et al.~\cite{zeng2008data} propose an automatic accident detection approach using D-S evidence theory data fusion based on the probabilistic output of multi-class SVMs. In~\cite{sadeky2010real}, it presents a real-time automatic traffic accidents detection method using Histogram of Flow Gradient (HFG) and the trajectory of vehicles by which the accident was occasioned is determined in case of occurrence of an accident. In~\cite{kamijo2000traffic}, it develops an extendable robust event recognition system for Traffic Monitoring and Accident Detection based on the hidden Markov model (HMM). ~\cite{chen2010automatic} proposed a method using SVM based on traffic flow measurement. A similar approach using BP-ANN for accident detection has been proposed in~\cite{srinivasan2004evaluation,ghosh2003wavelet}. In~\cite{saunier2010large}, it presented a refined probabilistic framework for the analysis of road-user interactions using the identification of potential collision points for estimating collision probabilities. Other methods for Traffic Accident Detection has also been presented using Matrix Approximation~\cite{xia2015vision}, optical flow and Scale Invariant Feature Transform (SIFT)~\cite{chen2016vision}, Smoothed Particles Hydrodynamics (SPH)~\cite{ullah2015traffic}, and adaptive traffic motion flow modeling~\cite{maaloul2017adaptive}. With advances in object detection with deep neural networks, several convolutional neural networks (CNNs) based automatic traffic accident detection methods~\cite{singh2018deep,sultani2018real} and recurrent neural networks (RNNs) based traffic accident anticipation methods~\cite{chan2016anticipating,suzuki2018anticipating} have been proposed along with some traffic accident dataset~\cite{sultani2018real,suzuki2018anticipating,kataoka2018drive,shah2018accident} of surveillance videos or dashcam videos~\cite{chan2016anticipating}. However, either most of these methods do not have real-time performances for online accident detection without using future frames, or most of these methods mentioned above give unsatisfactory results. Besides that, no proposed dataset contains videos with top-down views such as drone/Unmanned Aerial Vehicles (UAVs) videos, or omnidirectional camera videos for traffic analysis.
\begin{figure}
  \includegraphics[width=0.9\textwidth]{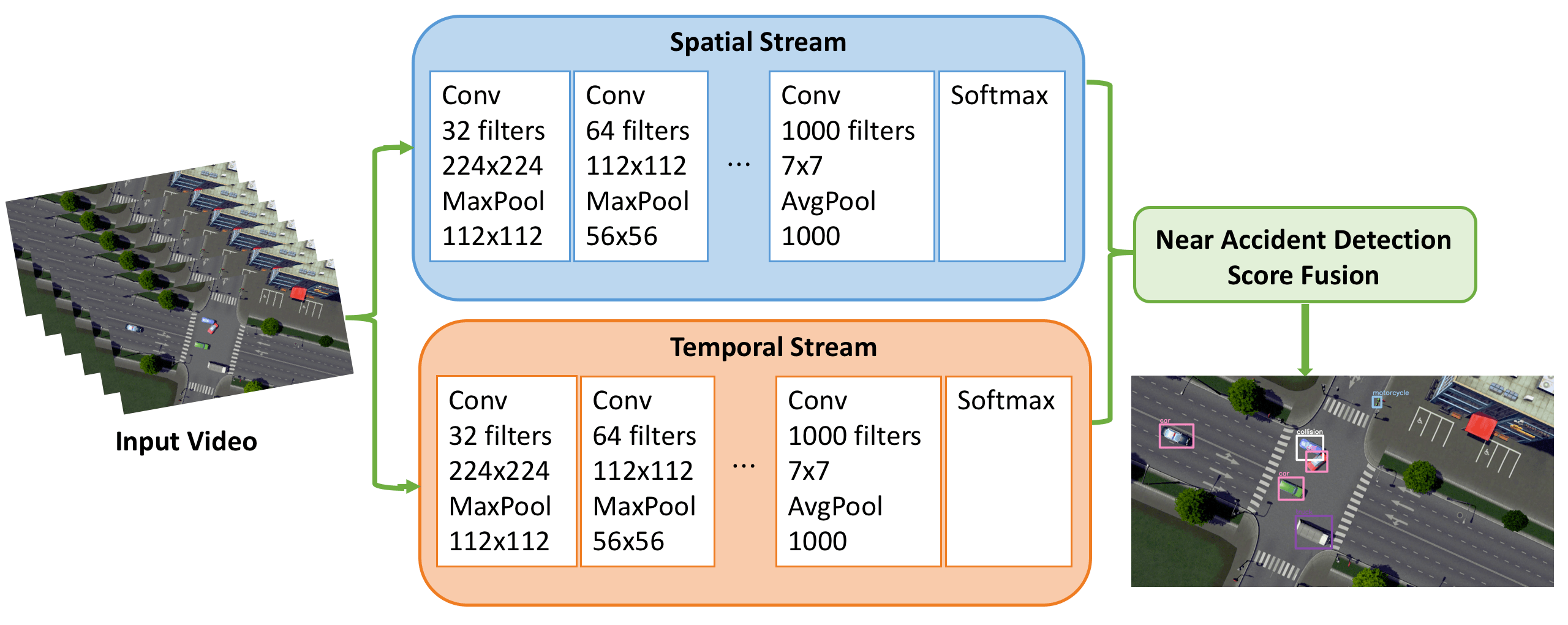}
  \caption{Two-stream architecture for near accident detection.}
  \label{fig:two}
\end{figure}
\section{Two-Stream architecture for Near Accident Detection}\label{sec:method}
We present our vision-based two-stream architecture for real-time near accident detection based on real-time object detection and multiple object tracking. The goal of near accident detection is to detect likely collision scenarios across video frames and report these near accident records. As videos can be decomposed into spatial and temporal components. We divide our framework into a two-stream architecture as shown in Fig~\ref{fig:two}. The spatial part consists of individual frame appearance information about scenes and objects in the video. The temporal part contains motion information for moving objects. For spatial stream convolutional neural network, we utilized a standard convolutional network of a state-of-the-art object detection method~\cite{redmon2016you} to detect individual vehicles and likely near accident regions at the single frame level. The temporal stream network is leveraging object candidates from object detection CNNs and integrates their appearance information with a fast multiple object tracking method to extract motion features and compute trajectories. When two trajectories of individual objects start intersecting or become closer than a certain threshold, we'll label the region covering two objects as high probability near accident regions. Finally, we take average near accident probability of spatial stream network and temporal stream network and report the near accident record.

\subsection{Preliminaries}
\begin{figure}
   \includegraphics[width=\textwidth]{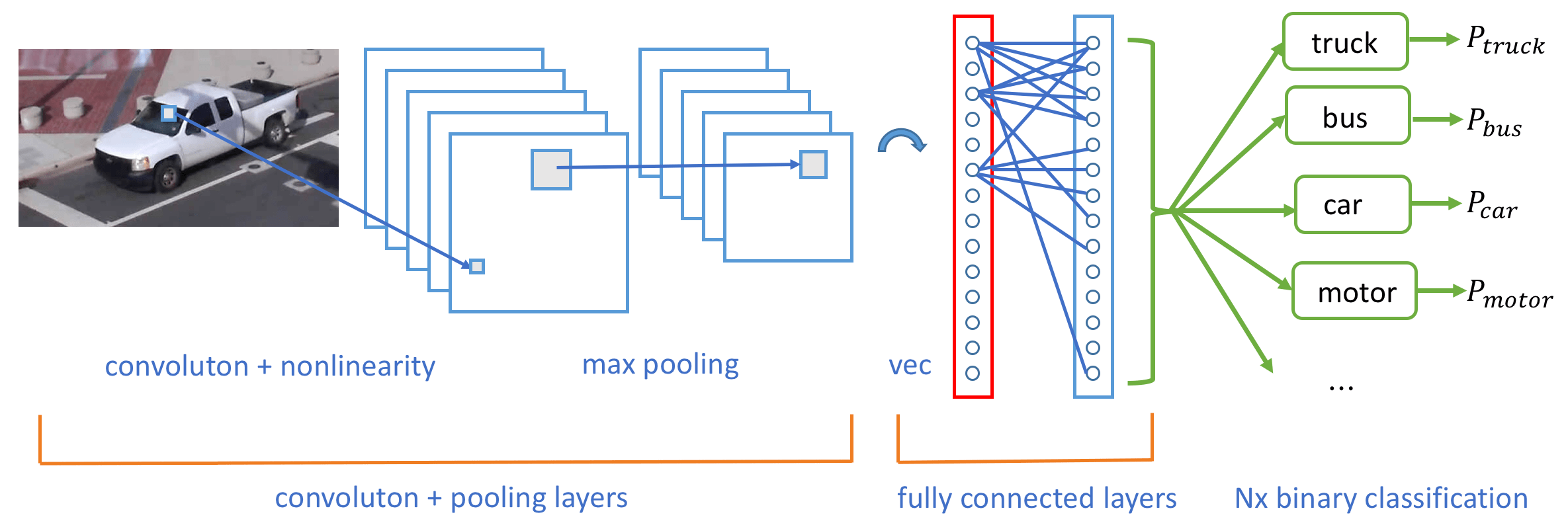}
   \caption[]{Architecture of Convolutional Neural Networks for image classification.\footnotemark}
  \label{fig:cnn}
\end{figure}

\textbf{Convolutional Neural Networks:} Convolutional Neural Networks (CNNs) have a strong capability of representing objects and helps to boost the performance of numerous vision tasks, comparing to traditional heuristic features \cite{dalal2005histograms}. A Convolutional Neural Networks (CNN) is a  a class of deep neural networks which is widely applied for visual imagery analysis in computer vision. A standard CNN usually consists of an input and an output layer, as well as multiple hidden layers (convolutional layers, pooling layers, fully connected layers and normalization layers) as shown in Figure~\ref{fig:cnn}. The input to a convolutional layer is an original image $\boldsymbol{X}$. We denote the feature map of $i$-th convolutional layer as $\boldsymbol{H}_i$ and $\boldsymbol{H}_0=\boldsymbol{X}$. Then $\boldsymbol{H}_i$ can be described as 
\begin{equation}
{\boldsymbol{H}_i} = f\left( {{\boldsymbol{H}_{i - 1}} \otimes {\boldsymbol{W}_i} + {\boldsymbol{b}_i}} \right)
\end{equation}
where $\boldsymbol{W}_i$ is the weight for $i$-th convolutional kernel, and $\otimes$ is the convolution operation of the kernel and $i-1$-th image or feature map. Output of convolution operation are summed with a bias $\boldsymbol{b}_i$. Then the feature map for $i$-th layer can be computed by applying a nonlinear activation function to it. Take an example of using a $32\times32$ RGB image with a simple ConvNet for CIFAR-10 classification~\cite{krizhevsky2009learning}.
\begin{itemize}
    \item Input layer: the original image with raw pixel values as width is 32, height is 32, and color channels (R,G,B) is 3.
    \item Convolutional layer: compute output of neurons which are connected to local regions in the image through activation functions. If we use 12 filters, we have result in volume such as $[32\times32\times12]$.
    \item Pooling layer: perform a downsampling operation, resulting in volume such as $[16\times16\times12]$.
    \item Fully connected layer: compute the class scores, resulting in volume of size $[1\times1\times10]$, where these 10 numbers are corresponding to 10 class score.
\end{itemize}
In this way, CNNs transform the original image into multiple high-level feature representations layer by layer and compute the final final class scores.

\footnotetext{The credit should be given to Adit Deshpande and his blog.}

\subsection{Spatial stream network}
In our framework, each stream is implemented using a deep convolutional neural network. Near accident scores are combined by the averaging score. Since our spatial stream ConvNet is essentially an object detection architecture, we build it upon the recent advances in object detection with YOLO detector~\cite{redmon2016you}, and pre-train the network from scratch on our dataset containing multi-scale drone, fisheye and simulation videos. As most of our videos contain traffic scenes with vehicles and traffic movement in top-down view, we specify different vehicle classes such as motorcycle, car, bus, and truck as object classes for training the detector. Additionally, near accident or collision can be detected from single still frame either from the beginning of a video or stopped vehicles associated in an accident after collision. Therefore, we  train our detector to localize these likely near accident scenarios. Since the static appearance is a useful cue, the spatial stream network effectively performs object detection by operating on individual video frames.

\subsubsection{YOLO object detection}
\begin{figure}
  \includegraphics[width=0.9\linewidth]{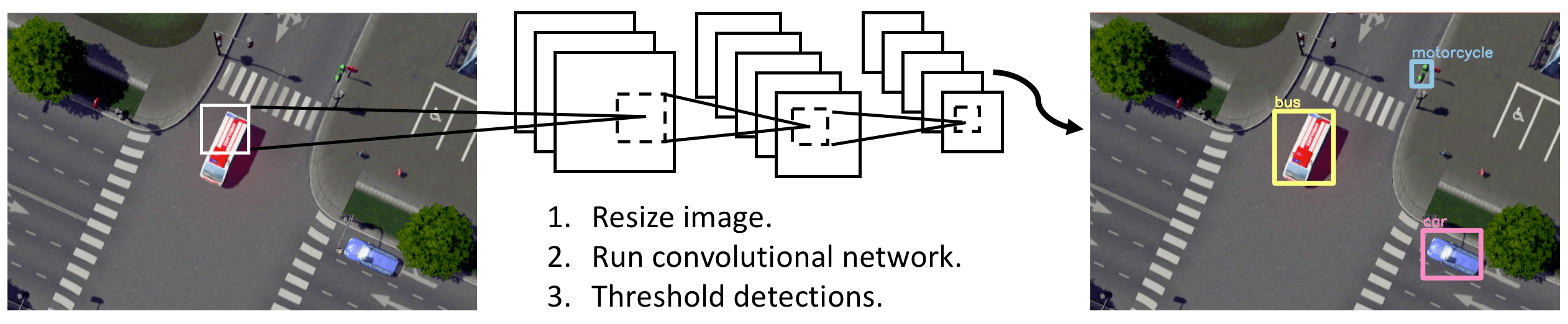}
  \caption{The YOLO Detection System~\cite{redmon2016you}. It (1) resizes the input image to $448 \times 448$, (2) runs a single convolutional network on the image, and (3) thresholds the resulting detections by the model's confidence.}
  \label{fig:2}
\end{figure}

You Only Look Once (YOLO)~\cite{redmon2016you} is a state-of-the-art, real-time object detection system. This end-to-end deep learning approach does not need region proposals and is much different from the region based algorithms. The pipeline of YOLO~\cite{redmon2016you} is pretty straightforward: YOLO~\cite{redmon2016you} passes the whole image through the neural network only once where the title comes from (You Only Look Once) and returns bounding boxes and class probabilities for predictions. Figure~\ref{fig:2} demonstrates the detection model and system of YOLO~\cite{redmon2016you}. In YOLO~\cite{redmon2016you}, it regards object detection as an end-to-end regression problem and uses a single convolutional network predicts the bounding boxes and the class probabilities for these boxes. It first takes the image and split it into an $S \times S$ grid, within each of the grid we take $m$ bounding boxes. For each grid cell,
\begin{itemize}
    \item it predicts B boundary boxes and each box has one box confidence score
    \item it detects one object only regardless of the number of boxes B
    \item it predicts C conditional class probabilities (one per class for the likeliness of the object class)
\end{itemize}

For each of the bounding box, the convolutional neural network (CNN) outputs a class probability and offset values for the bounding box. Then it selects bounding boxes which have the class probability above a threshold value and uses them to locate the object within the image. In detail, each boundary box contains 5 elements: $(x, y, w, h)$ and a box confidence. The $(x, y)$ are coordinates which represent the center
of the box relative to the bounds of the grid cell. The $(w,h)$ are width and height. These elements are normalized as $x$, $y$, $w$ and $h$ are all between 0 and 1. The confidence prediction represents the intersection over union (IoU) between the predicted box and any ground truth box which reflects how likely the box contains an object (objectness) and how accurate is the boundary box. The mathematical definitions of those scoring and probability terms are:
\begin{center}
    box confidence score $\equiv P_{r}(object)\cdot IoU$\\
    conditional class probability $\equiv P_{r}(class_{i}|object)$\\
    class confidence score $\equiv P_{r}(class_{i})\cdot IoU$\\
    class confidence score $=$ box confidence score $\times$ conditional class probability
\end{center}
where $\equiv P_{r}(object)$ is the probability the box contains an object. $IoU$ is the IoU between the predicted box and the ground truth. $\equiv P_{r}(class_{i})$ is the probability the object belongs to $class_{i}$. $\equiv P_{r}(class_{i}|object)$ is the probability the object belongs to $class_{i}$ given an object is presence. The network architecture of YOLO~\cite{redmon2016you} simply contains 24 convolutional layers followed by two fully connected layers, reminiscent of AlexNet and even earlier convolutional architectures. Some convolution layers use $1 \times 1$ reduction layers alternatively to reduce the depth of the features maps. For the last convolution layer, it outputs a tensor with shape $(7, 7, 1024)$ which is flattened. YOLO~\cite{redmon2016you} performs a linear regression using two fully connected layers to make boundary box predictions and to make a final prediction using threshold of box confidence scores. The final loss adds localization, confidence and classification losses together. 

\begin{table*}[h]
\begin{center}
\begin{tabular}{c|c|c|c}
Type & Filters & Size/Stride & Output\\
\hline
Convolutional & 32 & $3 \times 3$ & $224 \times 224 $ \\
Maxpool & &$2 \times 2 / 2$ & $112 \times 112 $ \\
Convolutional & 64 & $3 \times 3$ & $112 \times 112 $ \\
Maxpool & & $2 \times 2 / 2$ & $56 \times 56 $ \\
Convolutional & 128 &$3 \times 3$ & $56 \times 56 $ \\
Convolutional & 64 &$1 \times 1$ & $56 \times 56 $ \\
Convolutional & 128 &$3 \times 3$ & $56 \times 56 $ \\
Maxpool & & $2 \times 2 / 2$ & $28 \times 28 $ \\
Convolutional & 256 & $3 \times 3$ & $28 \times 28 $ \\
Convolutional & 128 & $1 \times 1$ & $28 \times 28 $ \\
Convolutional & 256& $3 \times 3$ & $28 \times 28 $ \\
Maxpool & & $2 \times 2 / 2$ & $14 \times 14 $ \\
Convolutional & 512 & $3 \times 3$ & $14 \times 14 $ \\
Convolutional & 256& $1 \times 1$ & $14 \times 14 $ \\
Convolutional & 512 & $3 \times 3$ & $14 \times 14$ \\
Convolutional & 256& $1 \times 1$ & $14 \times 14$ \\
Convolutional & 512 & $3 \times 3$ & $14 \times 14 $ \\
Maxpool & & $2 \times 2 / 2$ & $7 \times 7 $ \\
Convolutional & 1024 & $3 \times 3$ & $7 \times 7 $ \\
Convolutional & 512 & $1 \times 1$ & $7 \times 7 $ \\
Convolutional & 1024 & $3 \times 3$ & $7 \times 7$ \\
Convolutional & 512 & $1 \times 1$ & $7 \times 7$ \\
Convolutional & 1024 & $3 \times 3$ & $7 \times 7$ \\
\hline
\hline
Convolutional & 1000 & $1 \times 1$ & $7 \times 7$ \\
Avgpool & & Global & $1000$ \\
Softmax & & &\\
\end{tabular}
\end{center}
\caption{\small \textbf{Darknet-19~\cite{redmon2017yolo9000}.}}
\label{net}
\end{table*}

\scriptsize
\begin{multline}
\lambda_\textbf{coord}
\sum_{i = 0}^{S^2}
    \sum_{j = 0}^{B}
     \mathlarger{\mathbbm{1}}_{ij}^{\text{obj}}
            \left[
            \left(
                x_i - \hat{x}_i
            \right)^2 +
            \left(
                y_i - \hat{y}_i
            \right)^2
            \right]
+ \lambda_\textbf{coord} 
\sum_{i = 0}^{S^2}
    \sum_{j = 0}^{B}
         \mathlarger{\mathbbm{1}}_{ij}^{\text{obj}}
         \left[
        \left(
            \sqrt{w_i} - \sqrt{\hat{w}_i}
        \right)^2 +
        \left(
            \sqrt{h_i} - \sqrt{\hat{h}_i}
        \right)^2
        \right]
\\
+ \sum_{i = 0}^{S^2}
    \sum_{j = 0}^{B}
        \mathlarger{\mathbbm{1}}_{ij}^{\text{obj}}
        \left(
            C_i - \hat{C}_i
        \right)^2
+ \lambda_\textrm{noobj}
\sum_{i = 0}^{S^2}
    \sum_{j = 0}^{B}
    \mathlarger{\mathbbm{1}}_{ij}^{\text{noobj}}
        \left(
            C_i - \hat{C}_i
        \right)^2
+ \sum_{i = 0}^{S^2}
\mathlarger{\mathbbm{1}}_i^{\text{obj}}
    \sum_{c \in \textrm{classes}}
        \left(
            p_i(c) - \hat{p}_i(c)
        \right)^2
\end{multline}
\normalsize
where $\mathbbm{1}_i^{\text{obj}}$ denotes if object appears in cell $i$ and $\mathbbm{1}_{ij}^{\text{obj}}$ denotes that the $j$th bounding box predictor in cell $i$ is ``responsible'' for that prediction. YOLO~\cite{redmon2016you} is orders of magnitude faster (45 frames per second) than other object detection approaches which means it can process streaming video in realtime and achieves more than twice the mean average precision of other real-time systems. For the implementation, we leverage the extension of YOLO~\cite{redmon2016you}, Darknet-19, a classification model that used as the base of YOLOv2~\cite{redmon2017yolo9000}. The full network description of it is shown in Table~\ref{net}. Darknet-19~\cite{redmon2017yolo9000} has 19 convolutional layers and 5 maxpooling layers and it uses batch normalization to stabilize training, speed up convergence, and regularize the model~\cite{ioffe2015batch}.

\subsection{Temporal stream network}
The spatial stream network is not able to extract motion features and compute trajectories due to single-frame inputs. To leverage these useful information, we present our temporal stream network, a ConvNet model which performs a tracking-by-detection multiple object tracking algorithm~\cite{bewley2016simple,wojke2017simple} with data association metric combining deep appearance features. The inputs are identical to the spatial stream network using the original video.  Detected object candidates (only vehicle classes) are used to for tracking handling, state estimation, and frame-by-frame data association using SORT~\cite{bewley2016simple} and DeepSORT~\cite{wojke2017simple}, the real-time multiple object tracking methods. The multiple object tracking models each state of objects and describes the motion of objects across video frames. With tracking, we obtain results by stacking trajectories of moving objects between several consecutive frames which are useful cues for near accident detection. 

\subsubsection{SORT}
Simple Online Realtime Tracking (SORT)~\cite{bewley2016simple} is a simple, popular and fast Multiple Object Tracking (MOT) algorithm. The core idea is to perform a Kalman filtering~\cite{kalman1960new} in image space and do frame-by-frame data association using the Hungarian methods~\cite{kuhn1955hungarian} with an association metric that measures bounding box overlap. Despite only using a rudimentary combination of the Kalman Filter~\cite{kalman1960new} and Hungarian algorithm~\cite{kuhn1955hungarian} for the tracking components, SORT~\cite{bewley2016simple} achieves an accuracy comparable to state-of-the-art online trackers. Moreover, due to the simplicity of it, SORT~\cite{bewley2016simple} can updates at a rate of 260 Hz on single machine which is over 20x faster than other state-of-the-art trackers.

\textbf{Estimation Model.} The state of each target is modelled as:
\begin{equation}
    \mathbf{x} = [u,v,s,r,\dot{u},\dot{v},\dot{s}]^T,
\end{equation}
where $u$ and $v$ represent the horizontal and vertical pixel location of the centre of the target, while the scale $s$ and $r$ represent the scale (area) and the aspect ratio (usually considered to be constant) of the target's bounding box respectively. When a detection is associated to a target, it updates the target state using the detected bounding box where the velocity components are solved optimally via a Kalman filter framework~\cite{kalman1960new}. If no detection is associated to the target, its state is simply predicted without correction using the linear velocity model.

\textbf{Data Association.} In order to assign detections to existing targets, each target's bounding box geometry is estimated by predicting its new location in the current frame. The assignment cost matrix is defined as the IoU distance between each detection and all predicted bounding boxes from the existing targets. Then the assignment problem is solved optimally using the Hungarian algorithm~\cite{kuhn1955hungarian}. Additionally, a minimum IoU is imposed to reject assignments where the detection to target overlap is less than $IoU_{min}$. The IoU distances of the bounding boxes are found so as to handle short term occlusion caused by passing targets. 

\textbf{Creation and Deletion of Track Identities.} When new objects enter or old objects vanish in video frames, unique identities for objects need to be created or destroyed accordingly. For creating trackers, we consider any detection with an overlap less than $IoU_{min}$ to signify the existence of an untracked object. Then the new tracker undergoes a probationary period where the target needs to be associated with detections to accumulate enough evidence in order to prevent tracking of false positives. Tracks could be terminated if they are not detected for $T_{Lost}$ frames to prevent an unbounded growth in the number of trackers and localization errors caused by predictions over long durations without corrections from the detector. 

\subsubsection{DeepSORT}
DeepSORT~\cite{wojke2017simple} is an extension of SORT~\cite{bewley2016simple} which integrates appearance information through a pre-trained association metric to improve the performance of SORT~\cite{bewley2016simple}. It adopts a conventional single hypothesis tracking methodology with recursive Kalman filtering~\cite{kalman1960new} and frame-by-frame data association. DeepSORT~\cite{wojke2017simple} helps to solve a large number of identities switching problem in SORT~\cite{bewley2016simple} and it can track objects through longer periods of occlusions. During online application, it establishs measurement-to-track associations using nearest neighbor queries in visual appearance space. 

\textbf{Track Handling and State Estimation}. The track handling and state estimation using Kalman filtering~\cite{kalman1960new} is mostly identical to the SORT~\cite{bewley2016simple}. The tracking scenario is defined using eight dimensional state space~$(u, v, \gamma, h, \dot{x}, \dot{y}, \dot{\gamma}, \dot{h})$ that contains the bounding box center position $(u, v)$, aspect ratio $\gamma$, height $h$, and their respective velocities in image coordinates. It uses a standard Kalman filter~\cite{kalman1960new} with a constant velocity motion and linear observation model, where it takes the bounding coordinates~$(u, v, \gamma, h)$ as direct observations of the object state.

\textbf{Data Association}. To solve the frame-by-frame association problem between the predicted Kalman states and the newly arrived measurements, it uses the Hungarian algorithm~\cite{kuhn1955hungarian}. In formulation, it integrates both motion and appearance information through combination of two appropriate metrics. For motion information, the (squared) Mahalanobis distance between predicted Kalman states and newly arrived measurements is utilized:
\begin{equation}
    d^{(1)}(i,j)
    =
    (\bm{d}_j - \bm{y}_i)^{\bm{T}}(\bm{S}_i)^{-1}(\bm{d}_j - \bm{y}_i)
\end{equation}
where the projection of the $i$-th track distribution into
measurement space is ~$(\bm{y}_i, \bm{S}_i)$ and the $j$-th bounding box detection is $\bm{d}_j$. The second metric measures the smallest cosine distance between the~$i$-th track and~$j$-th detection in appearance space:
\begin{equation}
    d^{(2)}(i, j)
    = \min \left \{ 1 - {\bm{r}_j}^{T} \bm{r}^{(i)}_k | \bm{r}^{(i)}_k\in \mathcal{R}_i \right \}
\end{equation}
Then this association problem is built with combination of both metrics using a weighted sum where the influence of each metric on combined association cost can be controlled through hyperparameter $\lambda$.
\begin{equation}
    c_{i,j}
    =
    \lambda \, d^{(1)}(i, j) + (1 - \lambda) d^{(2)}(i, j)
\end{equation}

\textbf{Matching Cascade}. Rather than solving measurement-to-track associations in a global way, it adopts a matching cascade introduced in~\cite{wojke2017simple} to solve a series of subproblems. In some situation, when occlusion happens to a object for a longer period of time, the subsequent Kalman filter~\cite{kalman1960new} predictions would increase the uncertainty associated with the object location. In consequent, probability mass spreads out in state space and the observation likelihood becomes less peaked. Intuitively, the association metric should account for this spread of probability mass by increasing the measurement-to-track distance. Therefore, the matching cascade strategy gives priority to more frequently seen objects to encode the notion of probability spread in the association likelihood.

\subsection{Near Accident Detection}
\begin{figure}
  \includegraphics[width=0.9\linewidth]{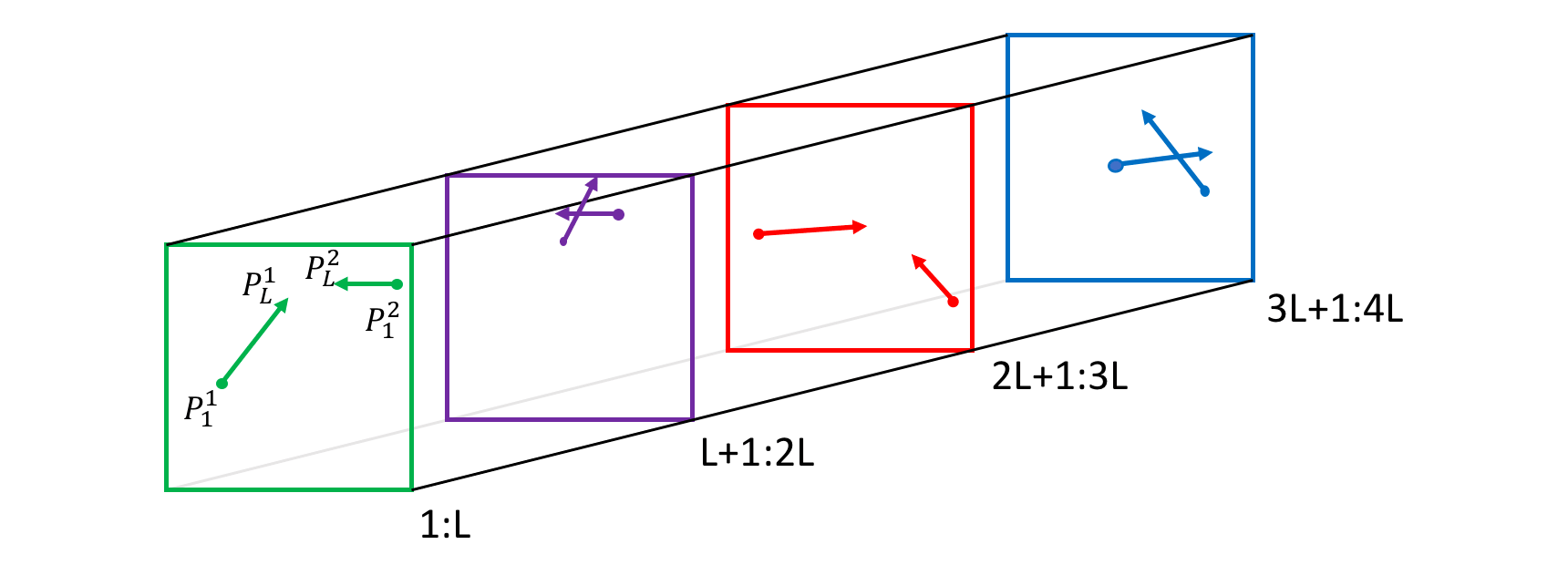}
  \caption{The stacking trajectories extracted from tracking. Consecutive frames and the corresponding displacement vectors are shown with the same colour.}
  \label{fig:traj}
\end{figure}

When utilizing the multiple object tracking algorithm,  we compute the center of each object in several consecutive frames to form stacking trajectories as our motion representation. These stacking trajectories can provide accumulated information through image frames, including the number of objects, their motion history and timing of their interactions such as near accident. We stack the trajectories of all object by every $L$ consecutive frames as illustrated in Figure~\ref{fig:traj} where $\mathbf{p}_t^i$ denotes the center position of the $i$-th object in the $t$-th frame. $\mathbf{P}_t = (\mathbf{p}_t^1, \mathbf{p}_t^2, ..., \mathbf{p}_t^{M_t})$ denotes trajectories of all the $M_t$ objects in the $t$-th frame. $\mathbf{P}_{1:t} = \{\mathbf{P}_1, \mathbf{P}_2, ..., \mathbf{P}_t\}$ denotes all the sequential trajectories of all the objects from the first frame to the $t$-th frame. As we only exam every $L$ consecutive frames, the stacking trajectories are sequentially as 
\begin{equation}
    \mathbf{P}_{1:L} = \{\mathbf{P}_1, \mathbf{P}_2, ..., \mathbf{P}_L\}, \mathbf{P}_{L+1:2L} = \{\mathbf{P}_{L+1}, \mathbf{P}_{L+2}, ..., \mathbf{P}_{2L}\},\cdots 
\end{equation}
$\mathbf{O}_t = (\mathbf{o}_t^1, \mathbf{o}_t^2, ..., \mathbf{o}_t^{M_t})$ denotes the collected observations for all the $M_t$ objects in the $t$-th frame. $\mathbf{O}_{1:t} = \{\mathbf{O}_1, \mathbf{O}_2, ..., \mathbf{O}_t\}$ denotes all the collected sequential observations of all the objects from the first frame to the $t$-th frame. We use a simple detection algorithm which finds collisions between simplified forms of the objects, using the center of bounding boxes. 

Our algorithm is depicted in Algorithm~\ref{alg:detect}. Once the collision is detected, we set the region covering collision associated objects to be a new bounding box with class probability of near accident to be 1. By averaging the near accident probability of output from spatial stream network and temporal stream network, we are able to compute the final outputs of near accident detection. 

\begin{algorithm}
 \KwIn{current frame $t_{current}$, collision state list $Collision$}
 \KwOut{collision state list $Collision$}
 \For{$t_{L} \leftarrow t_{previous}$ to $t_{current}$ in steps of $L$ frames}{
    \For{each pair of object trajectory ($\mathbf{p}_{:t_{L}}^1$, $\mathbf{p}_{:t_{L}}^2)$}{
    \If{($\mathbf{p}_{:t_{L}}^1$ intersects $\mathbf{p}_{:t_{L}}^2$ as of $t_{L}$)}{add $\mathbf{o}_1$, $\mathbf{o}_2$ to $Collision$}
    }
    \If{($Collisions$)}{$t_{previous} \leftarrow t_{L}$; 
return TRUE}
 }   
 $t_{previous} \leftarrow t_{d}$; return FALSE
\caption{Collision Detection}\label{alg:detect}
\end{algorithm}

\section{Experiments}\label{sec:experiments}
In this section, we first introduce our novel traffic near accident dataset (TNAD) and describe the preprocessing, implementation detail and experiments settings. Finally, we present qualitative and quantitative evaluation in terms of the performance of object detection, multiple object tracking, and near accident detection, and comparison between other methods and our framework.

\subsection{Traffic Near Accident Dataset (TNAD)}
\begin{figure}
  \includegraphics[width=1\linewidth]{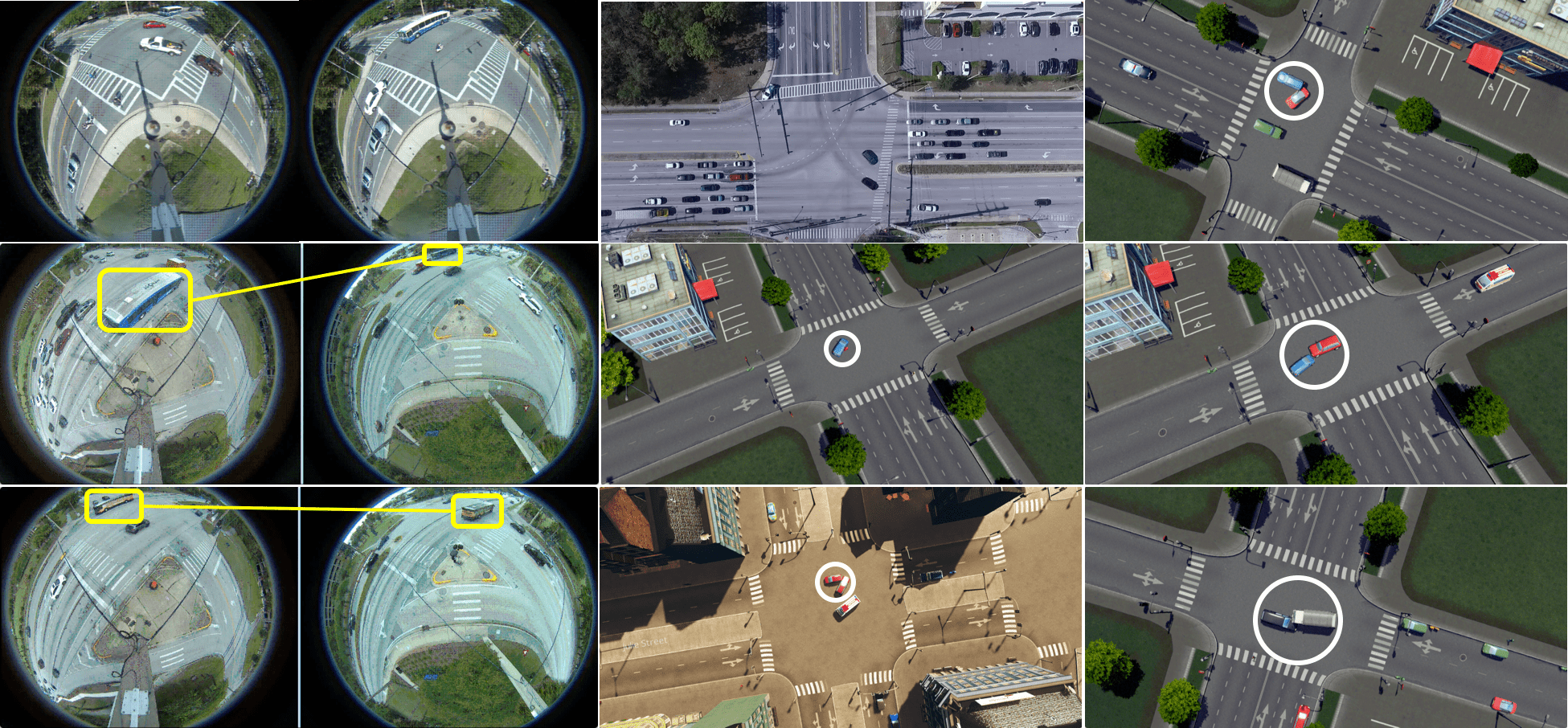}
  \caption{Samples of Traffic Near Accident Dataset (TNAD). Our dataset consists of a large number of diverse intersection surveillance videos and different near accident (cars and motorcycles). Yellow rectangles and lines represent the same object in multi-camera video. White circles represent the near accident regions.}
  \label{fig:dataset}
\end{figure}

As we mentioned in Section~\ref{sec:background}, there is no such a comprehensive traffic near accident dataset containing top-down views videos such as drone/Unmanned Aerial Vehicles (UAVs) videos, or omnidirectional camera videos for traffic analysis. Therefore, we have built our own dataset, traffic near accident dataset (TAND) which is depicted in Figure~\ref{fig:dataset}. Intersections tend to experience more and severe near accident due to factors such as angles and turning collisions. Traffic Near Accident Dataset (TNAD) containes 3 types of video data of traffic intersections that could be utilized for not only near accident detection but also other traffic surveillance tasks including turn movement counting. 

The first type is drone video that monitoring an intersection with top-down view. The second type of intersection videos is real traffic videos acquired by omnidirectional fisheye cameras that monitoring small or large intersections. It is widely used in transportation surveillance. These video data can be directly used as input for our vision-intelligent framework, and also pre-processing of fisheye correction can be applied to them for better surveillance performance. The third type of video is video data simulated by game engine for the purpose to train and test with more near accident samples. The traffic near accident dataset (TAND) consists of 106 videos with total duration over 75 minutes with frame rates between 20 fps to 50 fps.  The drone video and fisheye surveillance videos are recorded in Gainesville, Florida at several different intersections. Our videos are challenging than videos in other datasets due to the following reasons: 
\begin{itemize}
    \item Diverse intersection scene and camera perspectives: The intersections in drone video, fisheye surveillance video, and simulation video are much different. Additionally, the fisheye surveillance video has distortion and fusion technique is needed for multi-camera fisheye videos.
    \item Crowded intersection and small object: The number of moving cars and motorbikes per frame are large and these objects are relatively smaller than normal traffic video. 
    \item Diverse accidents: Accidents involving cars and motorbikes are all included in our dataset. 
    \item Diverse Lighting condition: Different lighting conditions such as daylight and sunset are included in our dataset.
\end{itemize}
We manually annotate the spatial location and temporal locations of near accidents and the still/moving objects with different vehicle class in each video. 32 videos with sparse sampling frames (only 20\% frames of these 32 videos are used for supervision) are used only for training the object detector. The remaining 74 videos are used for testing. 
\subsection{Fisheye and multi-camera video}
The fisheye surveillance videos are recorded from real traffic data in Gainesville.  We have collected 29 single-camera fisheye surveillance videos and 19 multi-camera fisheye surveillance videos monitoring a large intersection. We conduct two experiments, one directly using these raw videos as input for our system and another is first to do preprocessing for correcting fisheye distortion on video level and feed them into our system. As the original survellance video has many visual distortions especially near the circular boundaries of cameras, our system performs better on these after preprocessing videos. Therefore we keep the distortion correction preprocessing in the experiments for fisheye videos.

For large intersection, two fisheye cameras placed at opposite directions are used for surveillance and each of them mostly shows half of roads and real traffic for the large intersection.  In this paper, we do not investigate the real stitching problem (we'll leave it for further work). First, we do fisheye distortion correction and combine the two video with similar points. Then we apply a simple object level stitching methods by assigning the object identity for the same objects across the left and right video using similar features and appearing/vanishing positions. 

\subsection{Model Training}
The layer configuration of our spatial and temporal convolutional neural networks (based on Darknet-19~\cite{wojke2017simple}) is schematically shown in Table~\ref{net}. We adopt the Darknet-19~\cite{wojke2017simple} for classification and detection with deepSORT using data association metric combining deep appearance feature. We implement our framework on Tensorflow and do multi-scale training and testing with a single GPU (Nvidia Titan X Pascal). Training a single spatial convolutional network takes 1 day on our system with 1 Nvidia Titan X Pascal card. For classification and detection training, we use the same training strategy as YOLO9000~\cite{wojke2017simple}. We train the network on our TNAD dataset with 4 class of vehicle (motorcycle, bus, car, and truck) for 160 epochs using stochastic gradient descent with a starting learning rate of 0.1 for classification, and $10^{-3}$ for detection (dividing it by 10 at 60 and 90 epochs.), weight decay of 0.0005 and momentum of 0.9 using the Darknet neural network framework~\cite{wojke2017simple}.

\begin{figure}
  \includegraphics[width=1\linewidth]{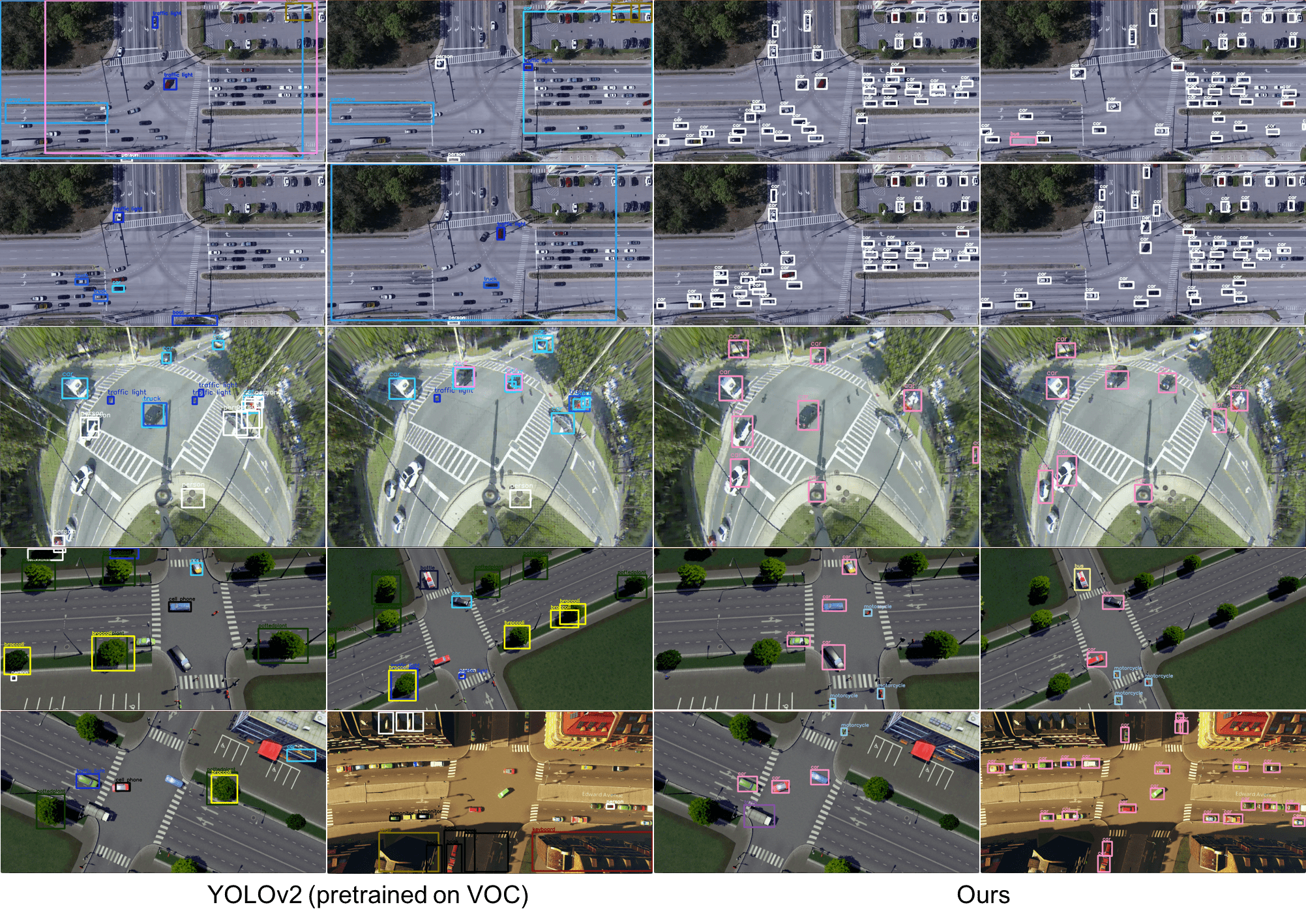}
  \caption{Sample results of object detection on TNAD dataset. \textbf{Left and middle left:} results of directly using YOLOv2~\cite{redmon2017yolo9000} detector pretrained on generic objects (VOC dataset)~\cite{Everingham15}. \textbf{Middle right and right:} results of our spatial network with multi-scale training based on YOLOv2~\cite{redmon2017yolo9000}.}
  \label{fig:objectdetection}
\end{figure}
\begin{figure}
  \includegraphics[width=\linewidth, height=8cm]{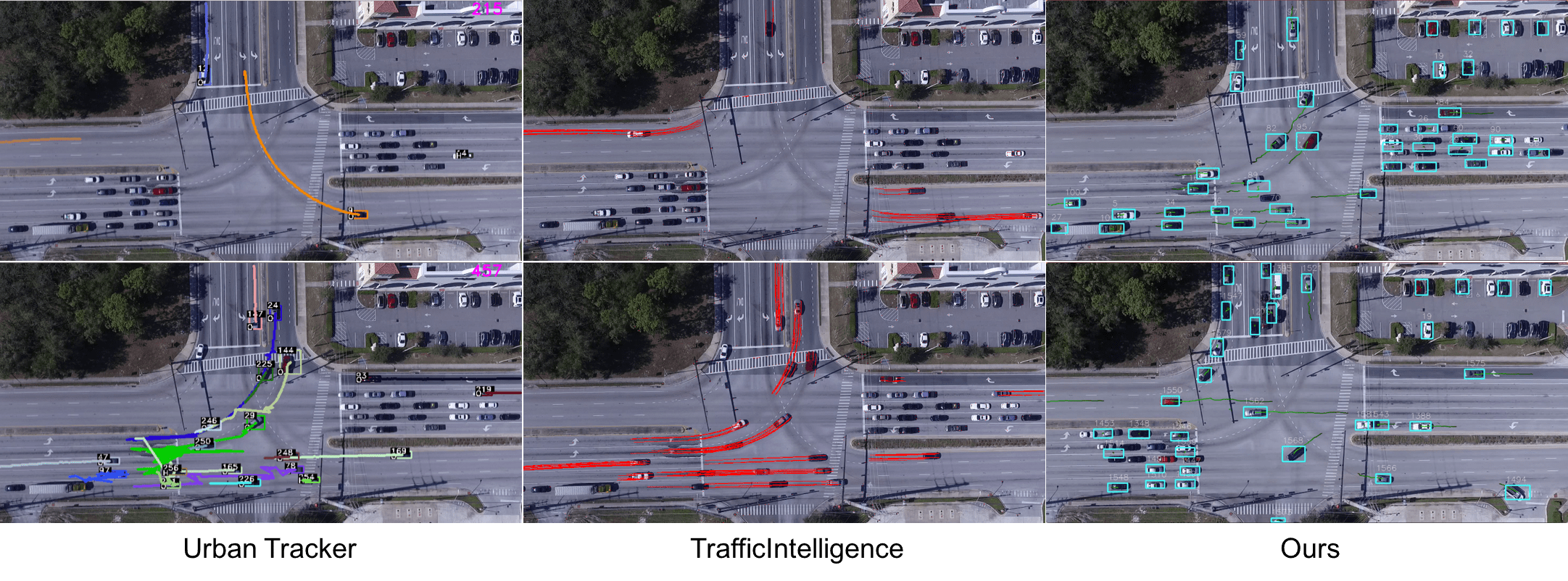}
  \caption{Tracking and trajectory comparison with Urban Tracker~\cite{jodoin2014urban} and TrafficIntelligence~\cite{jackson2013flexible} on drone videos of TNAD dataset.\textbf{Left:} results of Urban Tracker~\cite{jodoin2014urban}(BSG with Multilayer and Lobster Model). \textbf{Middle:} results of TrafficIntelligence~\cite{jackson2013flexible}. \textbf{Right:} results of our spatial network.}
  \label{fig:tracking}
\end{figure}
\begin{figure}
  \includegraphics[width=1\linewidth]{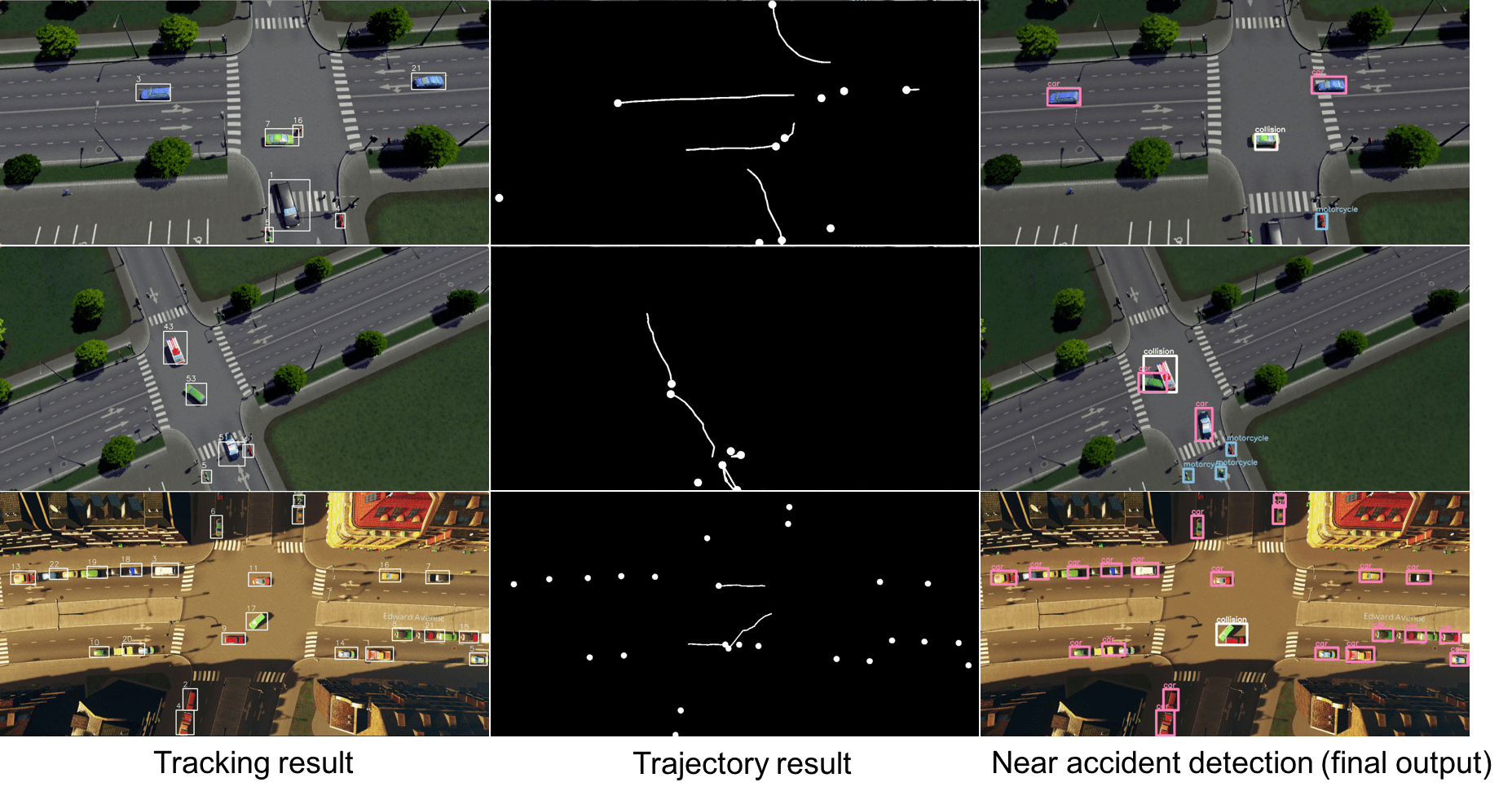}
  \caption{Sample results of tracking, trajectory and near accident detection of our two-stream Convolutional Networks on simulation videos of TNAD dataset.\textbf{Left:} tracking results based on DeepSORT~\cite{wojke2017simple}. \textbf{Middle:} trajectory results of our spatial network. ~\cite{jackson2013flexible}.\textbf{Right:} near accident detection results of our two-stream Convolutional Networks.}
  \label{fig:nearaccident}
\end{figure}

\begin{table}[htbp]
\caption{Frame level near accident detection results  }\label{tab:quantitative}
\resizebox{0.95\columnwidth}{!}{
\begin{tabular}{|c|c|c|c|c|}
\hline 
Video ID & near Accident (pos/neg) & \vtop{\hbox{\strut \# of frame for positive Near accident}\hbox{\strut  (groundtruth)/total frame}}   & \vtop{\hbox{\strut \# of frame for correct localization }\hbox{\strut (IoU >= 0.6)}}
  & \vtop{\hbox{\strut \# of frame for incorent localization}\hbox{\strut  (IoU<0.6)}}\tabularnewline
\hline 
\hline 
1 & pos & 12/245 & 12 & 0 \tabularnewline
\hline 
2 & neg &  0/259 & 0 &0 \tabularnewline
\hline 
3 & neg & 0/266 & 0 & 0 \tabularnewline
\hline 
4& pos &  16/267 & 13 & 0 \tabularnewline
\hline 
5 & pos & 6/246 & 4 & 0\tabularnewline
\hline 
6 & pos & 4/243 & 4 & 0\tabularnewline
\hline 
7 & neg & 0/286 & 0 & 0\tabularnewline
\hline 
8 & pos & 2/298 & 0 & 0\tabularnewline
\hline 
9 & pos & 27/351 & 23 & 6\tabularnewline
\hline 
10 & neg & 0/301 & 0  & 0\tabularnewline
\hline 
11 & neg & 0/294 & 0 & 0\tabularnewline
\hline 
12 & pos &  6/350 & 6 & 6 \tabularnewline
\hline 
13 & neg & 0/263 & 0 & 0 \tabularnewline
\hline 
14 & pos & 5/260 & 5 & 0 \tabularnewline
\hline 
15 & pos &  4/326  & 4 & 0 \tabularnewline
\hline 
16 & neg & 0/350  & 0  & 0 \tabularnewline
\hline 
17 & neg &  0/318  & 0 & 1\tabularnewline
\hline 
18 & pos & 10/340 & 8 & 0\tabularnewline
\hline 
19 & pos & 6/276  & 0 & 0\tabularnewline
\hline 
20 & pos & 8/428  & 4 & 0\tabularnewline
\hline 
21 & neg & 0/259  & 0 & 0 \tabularnewline
\hline 
22 & pos & 10/631 & 8 & 0 \tabularnewline
\hline 
23 & pos & 35/587  & 30 & 2 \tabularnewline
\hline 
24 & neg & 0/780  & 0 & 0 \tabularnewline
\hline 
25 & neg &  0/813 & 0 & 0\tabularnewline
\hline 
26 & neg &  0/765 & 0 & 0 \tabularnewline
\hline 
27 & pos &  8/616 & 8 & 0 \tabularnewline
\hline 
28 & pos & 10/243 & 10 & 1\tabularnewline
\hline 
29 & pos & 6/259  & 6 & 0 \tabularnewline
\hline 
30 & pos & 17/272  & 15 & 3 \tabularnewline
\hline 
\end{tabular}
}
\end{table}

\subsection{Qualitative results}

We present some example experimental results of object detection, multiple object tracking and near accident detection on our traffic near accident dataset (TNAD) for drone videos, fisheye videos, and simulation videos. For object detection (Figure~\ref{fig:objectdetection}), we present some detection results of directly using YOLO detector~\cite{redmon2017yolo9000} trained on generic objects (VOC dataset)~\cite{Everingham15} and results of our spatial network with multi-scale training based on YOLOv2~\cite{redmon2017yolo9000}. For multiple object tracking (Figure~\ref{fig:tracking}), we present comparison of our temporal network based on DeepSORT~\cite{wojke2017simple} with Urban Tracker~\cite{jodoin2014urban} and TrafficIntelligence~\cite{jackson2013flexible}. For near accident detection (Figure~\ref{fig:nearaccident}), we present near accident detection results along with tracking and trajectories using our two-stream Convolutional Networks method. The object detection results shows that, with multi-scale training on TNAD dataset, the performance of the detector significant improves. It can perform well vehicle detection on top-down view surveillance videos even for small objects. In addition, we can achieve fast detection rate at 20 to 30 frame per second. Overall, this demonstrates the effectiveness of our spatial neural network. For the tracking part, since we use a tracking-by-detection paradigm, our methods can handle still objects and measure their state where Urban Tracker~\cite{jodoin2014urban} and TrafficIntelligence~\cite{jackson2013flexible} can only handle tracking for moving objects. On the other hand, Urban Tracker~\cite{jodoin2014urban} and TrafficIntelligence~\cite{jackson2013flexible} can compute dense trajectories of moving objects with good accuracy but they have slower tracking speed around 1 frame per second. For accident detection, our two-stream Convolutional Networks are able to do spatial localization and temporal localization for diverse accidents regions involving cars and motorcycles. The three sub-tasks (object detection, multiple object tracking and near accident detection) can always achieve real-time performance at high frame rate, 20 to 30 frame per second according to the frame resolution (e.g. 28 fps for 960$\times$480 image frame).  Overall, the qualitative results demonstrate the effectiveness of our spatial neural network and temporal network respectively.


\begin{table}[]
\caption{Quantitative evaluation.}\label{tab:prerecall}
\begin{tabular}{|c|c|c|c|}
\hline
\multicolumn{2}{|c|}{\multirow{2}{*}{Benchmark Result}} & \multicolumn{2}{c|}{Predicted} \\ \cline{3-4} 
\multicolumn{2}{|c|}{}                            & Negative       & Positive      \\ \hline
\multirow{2}{*}{Actual}         & Negative        & 11081          & 19            \\ \cline{2-4} 
                                & Positive        & 32             & 160           \\ \hline
\end{tabular}
\end{table}

\subsection{Quantitative results}
Since our frame has three tasks and our dataset are much different than other object detection dataset, tracking dataset and near accident dataset such as dashcam accident dataset~\cite{chan2016anticipating}, it is difficult to compare individual quantitative performance for all three tasks with other methods. One of our motivation is to propose a vision-based solution for Intelligent Transportation System, we focus more on near accident detection and present quantitative analysis of our two-stream Convolutional Networks. The simulation videos are for the purpose to train and test with more near accident samples and we have 57 simulation videos with a total over 51,123 video frames. We sparsely sample only 1087 frames from them for whole training processing. We present the analysis of near accident detection for 30 testing videos (18 has positive near accident, 12 has negative near accident). Table~\ref{tab:quantitative} shows frame level near accident detection performance on 30 testing simulation videos. The performance of precision, recall and F-measure are presented in Table~\ref{tab:prerecall}. If a frame contains a near accident scenario and we can successfully localize it with Intersection of union (IoU) is large or equal than 0.6, this is a True Positive (TP). If we cannot localize it or localize it with Intersection of Union (IoU) is less than 0.6, this is a False Negative (FN). If a frame has no near accident scenario but we detect a near accident region, this is a False Positive (FP). Otherwise, this is a True Negative (TN). We compute the $\text{precision} = \frac{\text{TP}}{\text{TP + FP}}$, $\text{recall} = \frac{\text{TP}}{\text{TP + FN}}$, and F-measure $\text{F-measure} = \frac{2\times \text{precision} \times \text{recall}}{\text{precision + recall}}= \frac{\text{2TP}}{\text{2TP + FP + FN}}$. Our precision is about 0.894, recall is about 0.8333 and F1 score is about 0.863. The three sub-tasks (object detection, multiple object tracking and near accident detection) can always achieve real-time performance at high frame rate, 20~30 frame per second according to the frame resolution (e.g. 28 fps for 960$\times$480 image frame). In conclusion, we have demonstrated that our two-stream Convolutional Networks have an overall competitive performance for near accident detection on our TNAD dataset.

\section{Conclusion}\label{sec:conclusion}
We have proposed a two-stream Convolutional Network architecture that performs real-time detection, tracking, and near accident detection of road users in traffic video data. The two-stream Convolutional Networks consist of a spatial stream network to detect individual vehicles and likely near accident regions at the single frame level, by capturing appearance features with a state-of-the-art object detection method. The temporal stream network leverages motion features of detected candidates to perform multiple object Tracking and generate individual trajectories of each tracking target. We detect near accident by incorporating appearance features and motion features to compute probabilities of near accident candidate regions. We have present a challenging Traffic Near Accident dataset (TNAD), which contains different types of traffic interaction videos that can be used for several vision-based traffic analysis tasks. On the TNAD dataset, experiments have demonstrated the advantage of our framework with an overall competitive qualitative and quantitative performance at high frame rates. The future direction of the work is the image stitching mehtods for our proposed multi-camera fisheye videos.

\begin{acks}
The authors would like to thank City of Gainesville for providing real traffic fisheye video data.

\end{acks}

\bibliographystyle{ACM-Reference-Format}

\end{document}